\documentclass{IEEEcsmag}
\let\labelindent\relax
\usepackage[colorlinks,urlcolor=blue,linkcolor=blue,citecolor=blue]{hyperref}
\expandafter\def\expandafter\UrlBreaks\expandafter{\UrlBreaks\do\/\do\*\do\-\do\~\do\'\do\"\do\-}
\usepackage{upmath,color}

\jvol{XX}
\jnum{XX}
\paper{8}
\jmonth{Month}
\jname{Publication Name}
\jtitle{Publication Title}
\pubyear{2021}

\newcommand{\mask}[1]{\texttt{[MASK]} #1}
\newcommand{\relation}[1]{\texttt{[$r$]} #1}
\newcommand{\header}[1]{\texttt{[$v_h$]} #1}
\newcommand{\tailer}[1]{\texttt{[$v_t$]} #1}

\usepackage{enumitem}

\begin{document}

\sptitle{Theme Article: Graph Learning, Prompt Learning, Survey}

\title{A Survey of Graph Prompting Methods: Techniques, Applications, and Challenges}

\author{Xuansheng Wu$^{*}$}
\affil{University of Georgia, Athens, GA, 30605, USA}

\author{Kaixiong Zhou$^{*}$}
\affil{Rice University, Houston, TX, 77005, USA}

\author{Mingchen Sun}
\affil{Jilin University, Changchun, Jilin, 130015, China}

\author{Xin Wang}
\affil{Jilin University, Changchun, Jilin, 130015, China}

\author{Ninghao Liu}
\affil{University of Georgia, Athens, GA, 30605, USA}

\markboth{THEME/FEATURE/DEPARTMENT}{THEME/FEATURE/DEPARTMENT}

\begin{abstract}\looseness-1% 以下正好150词
The recent ``pre-train, prompt, predict'' training paradigm has gained popularity as a way to learn generalizable models with limited labeled data. The approach involves using a pre-trained model and a prompting function that applies a template to input samples, adding indicative context and reformulating target tasks as the pre-training task.
However, the design of prompts could be a challenging and time-consuming process in complex tasks. The limitation can be addressed by using graph data, as graphs serve as structured knowledge repositories by explicitly modeling the interaction between entities. In this survey, we review prompting methods from the graph perspective, where prompting functions are augmented with graph knowledge. In particular, we introduce the basic concepts of graph prompt learning, organize the existing work of designing graph prompting functions, and describe their applications and future challenges. This survey will bridge the gap between graphs and prompt design to facilitate future methodology development.
\end{abstract}

\maketitle
% \chapteri{G}raphs are ubiquitous in science and industry domains, such as social media networks, molecular graphs in biochemical informatics, knowledge graphs in natural language processing, and user-item interaction graphs in recommender systems.$^{1}$ 
% By modeling the interaction (edges) between entities (nodes), the graph data typically serves as structured knowledge repositories for information systems to understand complex associations between the nodes and infer new patterns. 
% For example, one might retrieve the neighborhood structures of target nodes for representation and interpretation; modern machine learning predicts the roles of nodes or edges, and infers the labels of whole graphs; large language models adopt external knowledge graphs to enhance reasoning capability.$^{30}$
\noindent Following the supervised training paradigm, the success of machine learning relies on a massive labeled dataset to learn the patterns between data samples and manual annotations for a specific task. However, %with a plethora of labeled data, 
it requires expensive costs %for humans 
to annotate the large dataset. 
% % 超字数
% Particularly, it is notorious that the supervised training suffers from poor generalization when annotations are limited or biased in the target space. 
The lack of labeled dataset has promoted researchers to investigate the ``pre-train, fine-tune'' framework.$^{2}$ 
The core idea is to leverage task-agnostic information to create pretext tasks for initializing the model, and then fine-tune it over the target task with less labeled samples. 
The knowledge learned from extensive pretext datasets improves model's generalization on the target problem. 
Under the ``pre-train, fine-tune'' framework, researchers focus on designing pretext task objectives and tailoring them to correlate with the target problem. 
Such engineering requires expert experience and time-consuming trials to find informative pretext objectives. 

Recently, prompt-based learning has enabled researchers to move away from expensive objective engineering towards simpler data engineering. 
Under the new training paradigm of ``pre-train, prompt, predict'', the prompt reformulates the target task looking similar to a pretext one to effectively reuse the pre-trained model.$^{3,4}$ 
Considering a triplet classification task with an example $<$Rose, Color, Red$>$, a prompting function could format it into the sentence ``The Color of Rose is \mask.'' and asks a language model to fill the masked blank with a color word. In this case, the language model pre-trained with predicting masked words, known as Masked Language Modeling (MLM) task,$^{5}$ could be applied directly to the reformulated problem. 
Such simplicity and efficiency have motivated the development of suitable prompting methods, including both manually and automatically designed prompt templates that concatenate input samples with language words or tokens. 

Early studies on prompt design often relied on expertise and trial-and-error to create intuitive templates for a broad range of target problems, which may inject noises and weaken generalization performance. 
Recently, some researches have suggested leveraging graph data (e.g., external knowledge graphs or task-dependent graphs) to induce precise contextual knowledge and differentiate between input samples. 
Graphs are ubiquitous in science and industry domains, such as social media networks, molecular graphs in biochemical informatics, knowledge graphs in natural language processing, and user-item interaction graphs in recommender systems.$^{1}$
The relational information serves as a knowledge base by storing features of nodes and edges and modeling their interactions, upon which one is easy to infer the related knowledge to fill in the prompt template. 
For example, based on Color's feature definition and Rose's neighborhood structure, one could reformulate a more indicative prompt as: ``What is the Color (with elementary colors of Red, Yellow, Blue) of Rose, including species of Darcey and Chrysler Imperial Roses? \mask''
This external knowledge from graph systems contains human experience or prior statistics to tailor the prompts for different input samples. 

In this survey, we review the rapidly growing area of prompt-based learning from a new perspective, where the prompts are generated upon graphs or designed for graph-related tasks.
We conduct a timely overview of state-of-the-art algorithms and applications in the field of graph-based prompt learning.
The intended audiences include machine learning researchers investigating how to leverage graph knowledge base to improve the design of prompt templates or applying the prompting methods to graph analysis. 
The contributions of this survey are summarized as follows:
\begin{itemize}
    \item[{\ieeeguilsinglright}] Compared with existing surveys that introduce general semantic-based prompts, we provide a unique overview of leveraging graph structures to inject adaptive knowledge into prompt design. To the best of our knowledge, this is the first review of graph prompting methods. 
    \item[{\ieeeguilsinglright}] We organize the graph prompting methods into two categories, including discrete prompt design and continuous prompt design, which differ in how to design prompt templates and reformulate the input samples.
    % % 超字数了
    % : (1) Manual prompt design creates informative templates based on human knowledge; (2)
    % Discrete prompt design injects the individual node and edge features centered on input samples; (3) Continuous prompt design applies differentiable graph representation learning methods and incorporates the node and edge embeddings in templates.
    \item[{\ieeeguilsinglright}] We summarize the state-of-the-art graph prompting methods for various applications, including graph machine learning, recommender systems, and natural language processing. We further discuss the limitations of existing pre-trained models and graph prompting methods, which sheds lights on the future researches.  % directions.   
\end{itemize}

\subsubsection{Comparison to Related Surveys.} 
There are limited surveys covering this newborn and growing rapidly topic. 
Liu et al. conducted the first review on prompt learning for natural language processing.$^{3}$  
They also provided an overview of prompt learning in recommendation systems.$^{4}$
In contrast, this paper focuses on reviewing studies that adopt the prompt learning method on graph data, no matter leveraging a graph knowledge base to improve the design of prompt templates or applying the prompting methods to graph analysis.

\begin{table*}[t]
\centering
\small
\label{tab:nota}
\caption{Notations in graph prompting methods.}
\begin{tabular}{llll}
\colrule
Name        & Notation & Example   \\ 
\toprule
\textit{Input}       &       $x$   &      Entity pair $<$Rose, Red$>$.   \\
\textit{Output}      &     $y$     &      Relation label $<$Color$>$.  \\ 
\textit{Pre-trained Model}      &     $f(\cdot)$     &  A pre-trained language model or graph model. \\ 
\toprule
\textit{Template} &      $ \mathcal{T}$    &   The relation between $[x_1]$ and $[x_2]$ is $\mask$.      \\
\textit{Prompt Addition}  &   $x^\prime = f_{\mathrm{prompt}}(x, \mathcal{T})$       &The relation between $<$Rose$>$ and $<$Red$>$ is $\mask$.      \\
\textit{Answer Set} & $\mathcal{A}$ & Possible answers $\{<$Color$>$,$<$Shape$>$,$<$Height$>\}$ to fill $\mask$. \\
\textit{Predicted Answer}  &    $\hat{z}$  & The predicted answer $<$Color$>$ to appear within the context.  \\
\botrule
\end{tabular}
\end{table*}

\section{Graph Learning: From Traditional Paradigms to Prompting Methods}
\subsection{Notations}
We denote sets with calligraphic capital letters (e.g., $\mathcal{D}$), matrices with boldface capital letters (e.g., $\mathbf{Z}$), vectors with boldface lowercase letters~(e.g., $\mathbf{v}$). 
We also denote a dataset as $\mathcal{D}=\{(x, y)\}$, where $x$ and $y$ are input sample and ground-truth information, respectively. 
% , and scalars or objects with lowercase alphabets~(e.g., $a_{ht}$ or $v$)
% the input sample $x$ could be any natural records, while the target label $y$ could be any human provided semantic labels. 
% Here, the space of input samples and target labels of the dataset are denoted as $\mathcal{X}$ and $\mathcal{Y}$, respectively.
% In this study, we focus on prompt learning on graphs, where the input sample $x$ and/or the output label $y$ are correlated with at least one kind of graph structure, such as a node instance, an edge, or a graph. 
In this work, we focus on summarizing the prompting function designs based on graphs or the prompting methods used in graph machine learning. Hence input sample $x$ can be instantiated as a sentence in NLP, a relational triplet in knowledge graph, a node of complex networks or biochemical graph. 
% Sometimes, there is a graph 
Let $G=(\mathcal{V}, \mathcal{E})$ denote the graph, %related to the dataset $\mathcal{D}$, 
where $\mathcal{V}$ and $\mathcal{E}$ are the sets of nodes and edges, respectively.
Each edge $e\in\mathcal{E}$ is described by a triplet $(v_h, r, v_t)$, $v_h, v_t\in\mathcal{V}$ are the head and tail nodes, and $r$ is the relation between the nodes. 
% In some cases, the relation could be ignored, such as in homogeneous graphs.
% We use $\mathbf{v}_h$, $\mathbf{v}_t$, and $\mathbf{r}$ to represent their high-dimensional embedding vectors, respectively. 
In vanilla case, the edge could be simplified as $e = (v_h, v_t)$, and the relation is represented by a scalar edge weight $a_{ht}\in\Re$. 

\subsection{Traditional Learning Paradigms}
% \paragraph{Supervised Learning}
% Traditionally, for a machine learning task with some graph-structured knowledge $G$, we collect a labeled dataset $\mathcal{D}=\{(x, y)\}$, and train a model $f$ using supervised learning to predict the label $\hat{y}=f(x;\theta)$ for matching the real label $y$ of samples in the dataset, where $\theta$ denotes the learnable model parameters. 
% Formally, the supervised training is formalized as:
% \begin{equation}
%     \min_{\theta}\, \sum_{(x,y)\in\mathcal{D}} \mathcal{L}(f(x; G, \theta); y), 
% \label{eq_sl}
% \end{equation}
% where $\mathcal{L}(\cdot;\cdot)$ is the loss function measuring the differences between predicted label $\hat{y}$ and real label $y$.

\paragraph{Supervised learning.} Typically, the supervised learning is formalized as: 
\begin{equation}
    \min_{\theta}\, \sum_{(x,y)\in\mathcal{D}} \mathcal{L}(f(x;  \theta); y),
\label{eq_sl}
\end{equation}
where $f(x;\theta)$ is model prediction based on input $x$ and learnable parameters $\theta$. $\mathcal{L}$ is loss function measuring the differences between the model prediction and ground truth $y$, such as cross-entropy or absolute difference loss. 

In NLP tasks, input $x$ is usually instantiated by text, and ground truth $y$ can be a discrete label or a textual tag. For example, in text classification, we take input $x =$ ``It is absolutely a great product.'' and generate label $y$ from $\{\mathrm{Positive, Negative}\}$. In text generation tasks, we are interested in question $x =$ ``How do you evaluate this item?'' and predict answer $y$ as ``Absolutely great''.

In graph learning tasks, input $x$ can be instantiated by a node, edge or graph. For example, the knowledge graph completion considers edge as input $x = <$Rose, Color, Red$>$ and predicts whether this relational triplet exists; the node/graph classification task takes a target node/graph %and its neighbors 
as $x$ and tries to predict the target label. 
% , and ground truth $y$ is a label or feature description

\paragraph{Pre-train and Fine-tune} 
Training a supervised model requires massive high-quality labeled data. 
However, manually labeling samples is expensive in many real-world scenarios. 
To tackle this challenge, the ``pre-train, fine-tune'' paradigm proposes to introduce a \emph{pre-training} stage to obtain transferable parameters of the model and then \emph{fine-tune} it on the target dataset. 

\subsubsection{Pre-training.}
The pre-training stage aims to learn generalizable knowledge from a massive task-agnostic dataset $\widetilde{\mathcal{D}}$ and avoids training from scratch given target tasks.
% given new tasks. 
% and obtain transferable parameters that can quickly adapt to downstream tasks, to avoid starting from scratch given new tasks. 
%According to the source of $\widetilde{\mathcal{D}}$, there are two types of pre-training tasks. 
%First, $\widetilde{\mathcal{D}}$ could be irrelevant to the graph structure $\mathcal{G}$, such as text corpus for pre-training language models.$^{5}$ 
%Second, $\widetilde{\mathcal{D}}$ could contain or even be the same as $\mathcal{G}$ for pre-training graph models.$^6$ 
% In this study, we focus on the dataset $\widetilde{\mathcal{D}}$ that contains $\mathcal{G}$ for pre-training graph models.$^6$ 
% The key to the success of pre-training $f_\mathrm{pre}$ is designing appropriate pretext tasks, which can be defined as follows:
The objective of pre-training is defined as:
\begin{equation}
    \min_{\theta}\, \sum_{(\widetilde{x}, \widetilde{y}) \in \widetilde{\mathcal{D}}}\mathcal{L}_{\mathrm{pre}}(f(\widetilde{x};\theta); \widetilde{y}),
\label{eq_pretrain}
\end{equation}
where $\mathcal{L}_{\mathrm{pre}}$ denotes the loss function of pretext task, $\widetilde{x}$ and $\widetilde{y}$ are the constructed input sample and label, respectively. The pre-training task is usually designed in a self-supervised manner to utilize the large amount of unlabeled data. For example, in pre-training language models, $\widetilde{x}$ can be a text with some tokens being masked, and $\widetilde{y}$ denotes those masked tokens to be recovered. In the graph learning tasks, $\widetilde{x}$ can be a subgraph of $G$ where some graph components (e.g., nodes or edges) are masked, and $\widetilde{y}$ denotes the masked components for reconstruction.

% Note that, $\widetilde{x}$ could be the same as the samples of the target dataset $x$.

\subsubsection{Fine-tuning.} 
The fine-tuning stage adapts pre-trained model $f$ % for a specific
on the target task according to Eq.~\eqref{eq_sl}.
% by using the downstream task dataset $\mathcal{D}$:
%Specifically, it develops a mapping module $f_\mathrm{tgt}$ on top with the pre-trained model $f_\mathrm{pre}$ to utilize the pre-training knowledge for the task: 
% \begin{equation}
%         \min_{\widetilde{\theta}}\, \sum_{(x,y)\in\mathcal{D}} \mathcal{L}(f(x;\widetilde{\theta}); y),
% %\min_{\theta_\mathrm{pre},\theta_\mathrm{tgt}}\, \sum_{(x,y)\in\mathcal{D}} \mathcal{L}(f_\mathrm{tgt}(f_{\mathrm{pre}}(x;\theta_\mathrm{pre}); \theta_\mathrm{tgt}); y),
% \end{equation} 
% where $\widetilde{\theta}$ is the pre-trained parameters optimized by Eq~\eqref{eq_pretrain}. 
The difference is model parameters $\theta$ are initialized based on the pre-training results. Practically, researchers could develop an additional architecture component (e.g., prediction head) on top of the pre-trained model to utilize the pre-training knowledge for the target problem.

\subsection{Graph Prompt Learning} 
One of the major issues in ``pre-train, fine-tune'' is the costly engineering of pretext task designs. If the pretext task is not necessarily related to the target problem (e.g., domain shifting), one may need to take more fine-tuning epochs to adapt model and even obtains poor generation performance. 
% On the other hand, the repeated model pre-training is required until an appropriate pretext task is figured out. 
To address the challenge, a new paradigm of ``pre-train, prompt, predict'' has been proposed to free the requirements of both massive supervised data and pretext task engineering. In particular, the prompt concept is introduced to augment the target input with more informative descriptions and reformulates it looking similar to the pre-training data. In this way, the pre-trained model can be directly reused to generate the desired output by behaving the same way as the pretext tasks. 
We mathematically describe the graph prompt learning in how to design the prompt template and leverage pre-trained model to conduct downstream tasks. 

% % 超字数了
% Although many existing works have designed the prompt templates based on intuitive human knowledge, the graph data provides a complementary way to better discover the informative context of target input for prompt design. In addition, the prompt concept has been extended to graph domain to improve the graph representation learning. 

% The ``pre-train, prompt, predict'' paradigm is another way to relax the labeled data requirement, enabling zero-shot learning (no labeled data).$^3$ To achieve this, it reformalizes the downstream task as one of the pre-training tasks, so that the pre-trained model can adapt to the downstream task by behaving the same way as the pre-training tasks. In particular, prompt learning on graphs consists of some additional modules to the pre-trained model $f_\mathrm{pre}$, including the prompting function $f_\mathrm{prompt}$ and additional steps that map the predictions of pre-trained models to the target labels. We introduce the steps of using prompts for model prediction including prompt addition, answer search and mapping, with an example in Table~1.
% \textcolor{red}{KX: A sentence to summarize the following three steps.}
% , i.e., $\mathcal{D}=\emptyset$

\subsubsection{Prompt Addition.} %  with Graphs
The prompting function $f_\mathrm{prompt}$ incorporates task-related knowledge $\theta_\mathrm{prompt}$ into input sample $x$ to generate a prompted input: 
\begin{equation}
\label{eq: prompt}
    x^\prime=f_\mathrm{prompt}(x;\theta_\mathrm{prompt}),
\end{equation} 
where $x^\prime$ lies on the same space as pre-training samples $\widetilde{x}$. Thus, the pre-trained model can take the prompted input $x^\prime$ to conduct pre-training tasks. 
%The data type of the task-related knowledge $\theta_\mathrm{prompt}$ could vary, such as being texts, continuous vectors, and parameters of additional graph-neural networks. 
Depending on the task-related knowledge parameters $\theta_\mathrm{prompt}$, the graph prompt can be categorized into two types: discrete and continuous ones. 
\begin{itemize}[leftmargin=*]
    \item The discrete prompt represents knowledge parameters $\theta_\mathrm{prompt}$ as the natural language words or individual nodes/edges, and puts them together with input data. 
% Generally, the task-related knowledge parameters $\theta_\mathrm{prompt}$ may contain one or more of the three forms, namely discriminate templates $\mathcal{T}=[w_1, ..., w_n]$, $d$-dimensional matrix $\mathbf{T}\in\Re$, and graph-structured data $G$. 
For example, considering the triplet completion task in knowledge graphs, % assume that we have a pre-trained language model as $f_\mathrm{pre}$, 
we could define $\theta_\mathrm{prompt}$ as a manually designed template $\mathcal{T}=$``The $\relation$ of {\texttt{[$v_h$]}} is $\mask$.'', where $\mask$ is the masked token to be filled by the pre-trained language model. 
Given an incomplete triplet input $x=$``$<$Rose, Color, [?]$>$'' without the tail node \tailer, we could generate the prompted input as $x^\prime=$``The Color of Rose is $\mask$.'' Template $\mathcal{T}$ could be expanded by concatenating the neighbors of input triplets based on the underlying graph database, e.g., $\mathcal{T}=$``The $\relation$ of $\header$ is $\mask$, where \header is related to {\texttt{[NEIGHBORS]}}.''
For example, ``The Color of Rose is $\mask$, where Rose is related to species of Darcey and Chrysler Imperial.'' 

\item The continuous prompt uses the differential vectors to represent knowledge parameters $\theta_\mathrm{prompt}$, which are prepended to input data in the continuous embedding space. Compared with the discrete version, the continuous prompt relaxes the constraint of human-interpretable semantic words or physical neighbors, and allows the prompt related parameters $\theta_\mathrm{prompt}$ to be optimized in an end-to-end manner. Based on the graph database, the continuous prompt could leverage the graph representation learning approaches, like graph neural networks, to obtain the structure-aware knowledge parameters. 
% % 超字数了
% Recalling the previous example, let $\bf{x}$ denote the embedding vector of incomplete triplet $x$, and let $\bf{T} = [\bf{t}_1, \cdots, \bf{t}_n]\in\Re^{n\times d}$ denote a sequence of input-specific neighborhood embeddings (e.g., those of Darcey and Chrysler Imperial Roses). The continuous prompt can be generated by simply concatenating them as:  $x^\prime= [\bf{x}; \bf{T}]$.
\end{itemize}

% Then, we can ask the pre-trained language model to fill with blank [MASK] with a color word to accomplish our triplet completion task. 
% In this specific example, we ignore side knowledge $\mathcal{G}$. 

\subsubsection{Answer Search and Mapping.} 
Based on the input prompt, the pre-trained model is leveraged to generate the desired output. Typically, we define an answer set $\mathcal{Z}$ containing all the possible target labels. We then search over $\mathcal{Z}$ to generate the desired output maximizing the possibility or similarity function $P$:
\begin{equation}
    z = \text{argmax}_{z^\prime\in\mathcal{Z}} P(f(x^\prime), z^\prime).
\end{equation} 

In NLP or knowledge graph completion tasks, where an language model is adopted to predict the sample classes, answer set $\mathcal{Z}$ could be a small subset of language words. Each candidate answer $z^\prime$ fills the masked prompt and estimates the corresponding possibility. For example, in the case of triplet completion for color prediction, we can define $\mathcal{Z} = \{\text{``Red''}, \text{``Blue''}, \text{``Yellow''}\}$ as the possible colors that any instance from the knowledge graph can have. In the product review classification problem, $\mathcal{Z} = \{\mathrm{Positive, Negative}\}$ are mapped to the binary ratings.

In the general graph learning tasks where the data samples are not associated with language description, answer set $\mathcal{Z}$ can be defined as a set of trainable class vectors. We measure similarity between the output $f(x^\prime)$ of pre-trained model (e.g., node representation vectors of GNNs) and the class vectors to obtain the target labels.

% We predict task label $\hat{y}$ by mapping the output of the pre-trained model from the pre-training label set 
% $\widetilde{\mathcal{Y}}$ to the task label set $\mathcal{Y}$. 
% Typically, we first define an answer set $\mathcal{A}\subseteq\widetilde{\mathcal{Y}}$ containing some pre-training labels that related to the task labels. 
% In the case of triplet completion for color prediction, we can define $\mathcal{A} = \{\text{``Red''}, \text{``Blue''}, \text{``Yellow''}, \text{``Green''}\}$ as the possible colors that any instance from the knowledge graph can have. 
% We then search the answer $\hat{z}\in\mathcal{A}$ having the highest probability:
% \begin{equation}
%     \hat{z} = \text{search}_{z\in\mathcal{A}} P(z^\prime=z),
% \end{equation} 
% where $z^\prime=f_\mathrm{pre}(x^\prime;\widetilde{\theta})$ is the prediction from the pre-trained model for the prompted input , and the search function could be \emph{argmax} or \emph{sampling} operations.
% In most cases, we can directly use the predicted answer $\hat{z}$ as the predicted output $\hat{y}$ for the target task, such as considering the answer $\hat{z}=\text{``Red''}$ as the predicted tail node $v_t$ in our triplet completion example. 

% However, this procedure may become more complex in some other tasks such as classification, where the each target label $y$ associates with multiple answers $z$. 

\subsection{Graph Prompt Design}
Graph prompting methods rely heavily on the prompting function as defined in Eq.~\eqref{eq: prompt}. In the rest of this paper, we systematically review the existing approaches in two categories: discrete and continuous graph prompts. Within each category, we further introduce the graph prompt from three aspects, namely \textit{manual prompt}, \emph{node-level prompt}, and \emph{topology-level prompt}. More specifically, the manual prompt constructs the template according to intuitive human knowledge and applies them to graph applications. While the node-level prompt utilizes the knowledge of a single node (e.g., node attributes and types), the topology-level prompt uses the relationships among multiple nodes (e.g., motifs) to formulate the graph prompt.

\begin{figure}
\vspace{-1cm}
\centerline{\includegraphics[width=0.99\linewidth]{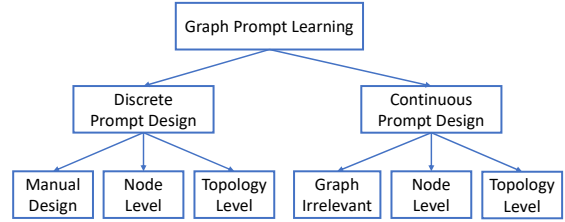}}
\vspace{-1.2cm}
\caption{A taxonomy of graph prompt learning techniques.}
\label{fig:taxonomy}
\vspace{-0.3cm}
\end{figure}

\section{Discrete Graph Prompt}
\label{discrete_prompt}
The design of prompting function could be motivated by graphs. The graph-induced discrete prompting function $f_{\mathrm{disc}}$ constructs a description of the input graph to introduce graph knowledge into prompts. 
Specifically, it generates graph-related textual knowledge and align them with manual textual templates to enrich the input.
Formally, the graph-induced discrete prompting function is given by:
\begin{equation}
    \label{eq: struc_prompt}
    x^\prime = f_{\mathrm{prompt}}(x; G, \mathcal{T}),
\end{equation}
where $x$ has a corresponding node $v$ from the graph $G$, and $\theta_\mathrm{prompt} = \{G, \mathcal{T}\}$ are the parameters of the prompting function.
The prompting function could either be designed manually or induced by the graph $G$.
The latter can be further divided into two categories, according to the type of graph knowledge they explored, namely node information and graph topology.
% Different from manual templates, the graph-induced discrete prompting function is proposed for leveraging the graph $G$ as side information for downstream tasks.
% That is, such discrete prompts both describe the downstream tasks and infuse domain knowledge from the graphs.

\subsection{Manual Prompt} % Manual Prompt Design
\label{manual_prompt}
A natural way to obtain a discrete template $\mathcal{T}$ is manually designing the template regardless of the graph. 
The manual prompting function creates prompted sample according to the human knowledge in downstream tasks. 
Formally, the prompting function is defined as: 
\begin{equation}
    \label{eq: fix_prompt}
     x^\prime=f_\mathrm{prompt}(x; \mathcal{T}),
\end{equation}
where $\mathcal{T}$ is the pre-defined prompt template and $f_\mathrm{prompt}$ inserts the input $x$ to each slot from the template $\mathcal{T}$.

The concrete design of manual template $\mathcal{T}$ could vary in different tasks or applications, where the key is to reformat the downstream task as one of the pre-training tasks. 
%For example, in recommendation tasks, P5 manually design the template as $\mathcal{T}=$``What star rating do you think [USER] will give [ITEM]? [MASK]'', where [USER] and [ITEM] are slots for filling the input user-item pair.$^7$ 
For example, in recommender systems, where the model is designed for predicting the user's preferences towards items, the P5 model manually design the prompt template as $\mathcal{T}=$``What star rating do you think \texttt{[USER]} will give to \texttt{[ITEM]}? $\mask$'', where \texttt{[USER]} and \texttt{[ITEM]} are slots for filling the input user-item pair.$^7$ 
Since this manual template shows that the task is a rating prediction, we can expect a pre-trained language model to conduct this task by filling $\mask$ with a number. 
In knowledge extraction tasks, RelationPrompt applies prompt learning for knowledge extraction by manually design a template as $\mathcal{T}=$``Context: \texttt{[x]}. Head Entity: $\mask$. Tail Entity: $\mask$. Relation: $\mask$.'', where \texttt{[x]} is the slot for the input text.$^8$ 
Given an input text $x=$``Their grandson was Captain Nicolas'', RelationPrompt expect a pre-trained language model can fill the three $\mask$ slots with ``Nicolas'', ``Captain'', and ``Military'', respectively. 
For the biomedical entity normalization task, GraphPrompt fills an input entity $x$ into a human designed template $\mathcal{T}=$``$\mask$ is identical with $[x]$'' to predict the synonym entity $\hat{y}$ of the input $x$.$^9$ 
% % 字数不够了
% Assuming that the input entity $x=$``CD115 (human)'', the prompted input could be $x^\prime=$``$\mask$ is identical with CD115 (human)'', and we expect the pre-trained language model to fill the $\mask$ with the synonym entity ``macrophage colony-stimulating factor 1 receptor (human)''. 

\subsection{Node-level Prompt} % Node Level Information
The adopted node information could vary for different tasks, but it typically describes the properties of vertices.$^{10}$ A key step in the existing studies is verbalizing the node information. Thus, the discrete prompting function is written as:
\begin{equation}
\begin{aligned}
   x^\prime = f_\mathrm{prompt}(x; g_{\mathrm{verb}}(x;G), \mathcal{T}), 
\end{aligned}
\end{equation}
where 
%$g_{\mathrm{match}}(x, G)$ is a alignment function matching the input sample $x$ to a node $v$ from the given graph $G$, 
$g_{\mathrm{verb}}$ is a non-parametric function describing the node attributes correlated to the input $x$ with natural language, 
and the prompting function $f_\mathrm{prompt}$ fills the template $\mathcal{T}$ with input $x$ and the verbalized attributes. Some concrete examples are provided below.

In knowledge graph completion tasks, where the node set $\mathcal{V}$ is a named entity set, the edge set $\mathcal{E}$ is the entity relation set, and a piece of knowledge is represented as a triplet $(u, e, v)$, $u,v\in\mathcal{V}$, and $e\in\mathcal{E}$. 
The goal is to judge whether a triplet $(u,e,v)$ is valid by converting it and its supporting information into textual sentences,$^{12}$  which will be fed into pre-trained language models for classification later.
Specifically, the attributes of the nodes $u$ and $v$ are the supporting information of the triplet, and the verbalizing function $g_\mathrm{verb}$ formats these information into sentences.
For example, given a node $u=$\emph{\underline{Lebron James}}, the verbalizing function $g_\mathrm{verb}$ returns the supportiing information as: \emph{``\underline{Lebron James}: American basketball player.''}

%We use recommender systems as an example to illustrate the idea of building prompts from node attributes.
In generative tasks, prompts could be formulated to ask for information from models. For example, in explainable recommender systems, the graph nodes include users $\mathcal{U}$ and items $\mathcal{I}$, and the edge set $\mathcal{E}$ shows user-item interactions.
PEPLER suggests incorporating user and item attributes as prompts to generate textual explanations,$^{11}$ where the verbalizing function $g_{\mathrm{verb}}$ is implemented as a retrieval function returning key features related to both the user $u\in\mathcal{U}$ and item $i\in\mathcal{I}$. 
% In this study, user and item features are words commonly mentioned by the reviews from that user or the reviews to that item. 
%One example of the prompted input of a user-item pair is ``dinner restaurant service [MASK]'', and a pre-trained language model is asked to generate a piece of text at [MASK] to explain the user preference to the item in terms of ``dinner'', ``restaurant'', and ``service'', such as ``The restaurant of this hotel provided me a wonderful dinner service.''. 
To be specific, if the commonly mentioned words by a user are \{``bed'', ``restaurant'', ``service''\} and the commonly mentioned words to a hotel are \{``location'', ``restaurant'', ``service''\}, the prompted input of this user-item pair would be \emph{``restaurant service } $\mask$ \emph{''}. 
Then a pre-trained language model is asked to generate a piece of text at $\mask$ to explain the user preference to the item in terms of ``restaurant'' and ``service'', such as ``The restaurant of this hotel provided me a wonderful dinner service.''

\subsection{Topology-level Prompt} % Topology Level Information
Compared with node attributes that describe the local information of graph components, topological structures of graphs carry broader and more diverse knowledge. 
Graph topology structures elicit specific prompts that not only focus on describing the edges and neighbors within a sub-graph, but also enable wider perspectives on sampling a sub-graph when provided with a sample.
%Discrete prompts induced by graph topology structures not only narrow on verbalizing the edges and neighbors within the sub-graph, but also broad visions on sampling a sub-graph by given a sample. 
The corresponding prompting function can be defined as:
\begin{equation}
\label{discre_graph_topology}
\begin{aligned}
    x^\prime &= f_\mathrm{prompt}(x; g_{\mathrm{verb}}(\widetilde{G}), \mathcal{T}),  \\
    \widetilde{G}&=g_{\mathrm{sample}}(x; G),\\
\end{aligned}
\end{equation}
where $g_{\mathrm{sample}}$ selects a sub-graph $\widetilde{G}\subset G$ around the center node correlated to the input $x$, and $g_{\mathrm{verb}}$ is a non-parametric function describing the sub-graph $\widetilde{G}$ in natural language words. We introduce different graph sampling methods for designing the prompting function as follows, including edge sampling, neighborhood sampling, and path sampling.
% \textcolor{red}{KX: In the following example, I am not sure how $\mathcal{T}$ works, and how to concatenate the verb results.}

Edges are fundamental graph components to augment prompting functions.
For example, in event causality detection, after mapping input tokens into node entities in a KG, a set of triples can be extracted from the KG to provide common-sense knowledge to describe events.$^{17}$  
In relation extraction, KGs provide rich ontology information of entities to help infer their relations.$^{15}$ 
Taking the input \emph{\underline{Bill Gates, co-founder of Microsoft.}} as an example, OntoPrompt first matches two named entity \emph{Bill Gates} and \emph{Microsoft} with ``person'' and ``organization'' on the KG, respectively, then identifies them as      ``entrepreneur'' and ``company'' according to KG ontology information, and finally retrieves  ``leader of'' as the relation to fill the prompt via the verbalizing function $g_{\mathrm{verb}}$.
% Specifically, OntoPrompt extends the definition of the node set $\mathcal{V}$ from a named entity set to an ontology set, and the edge set $\mathcal{E}$ is limited to relations among ontologies and those between named entities and ontologies.
% The sampling function $g_{\mathrm{sample}}$ extracts the connected nodes to the given named entity input $v$, and the verbalizing function $g_{\mathrm{verb}}$ formats the connected nodes into textual descriptions. 
% Taking the input \emph{\underline{Bill Gates, co-founder of Microsoft.}} as an example, OntoPrompt first matches two named entity \emph{Bill Gates} and \emph{Microsoft} on the knowledge graph. Then, the $g_{\mathrm{sample}}$ extracts 2-hops (ontology) neighbors of the two entities. 
% Finally, the verbalizing function $g_{\mathrm{verb}}$ generates verbal definitions of the ontology nodes as prompts.

Neighborhood sampling facilitates prompt formulation in providing context information of the target node. For example, in sequential recommender systems, by inducing user shopping history as graph topological information to obtain prompts. 
LMRecSys makes the first attempt to exploit the pre-trained language models for movie recommendation,$^{13}$ where the sampling function $g_{\mathrm{sample}}$ selects a few of the latest watched movies of the user, and the verbalizing function $g_{\mathrm{verb}}$ uses the movie titles as the descriptions of these selected movies. 
Given a user to the LMRecSys, the prompting function returns \emph{``A user watched \underline{Raiders of the Lost Ark}, \underline{Star Wars: Ep VI-Return of the Jedi}, \underline{Ran}. The user will also watch $\mask$.''}, 
%where the first sentence is generated by the verbalizing and sampling functions, and the second sentence is the template $\mathcal{T}$.
where the template $\mathcal{T}=$\emph{``A user watched }[$x_1$]\emph{, }[$x_2$]\emph{, }[$x_3$]\emph{. The user will also watch }$\mask$\emph{.''}, and the movie names are generated by the verbalizing and sampling functions.
Besides this study, M6-Rec designs a fine-grained verbalizing function $g_{\mathrm{verb}}$ to describe detailed features of the user-item interactions,$^{14}$ such as the category of the clicked movie and when did the user click each movie. 
One example of the prompted input from M6-Rec could be \emph{``A user watched \underline{Raiders of the Lost Ark} 14 days ago, \underline{Star Wars: Ep VI-Return of the Jedi} 4 days ago, \underline{Ran} 15 minutes ago. The user will also watch $\mask$.''}

% In the implicit event arguments extraction task, CUP proposes a multi-stages pipeline to utilize the topology Abstract Meaning Representation (AMR) graph.$^{16}$ 
% %Given a document, a pre-trained AMR parser constructs an AMR graph containing all extracted named entities $\mathcal{N}$ and their dependent relations $\mathcal{E}$. 
% Given a document and its AMR graph, CUP runs over a multi-stages pipeline according to the number of hops from the given trigger word to a target argument. This multi-stages pipeline actually is the sampling function $g_{\mathrm{sample}}$, while the verbalizing function $g_\mathrm{verb}$ is defined as extracting sentences by giving a sub-AMR graph. 
% %The language model is given increasing amounts of context at each stage, starting with just the sentence containing the trigger word and ending with the entire document. 
% The extracted attributes at each stage are used to update a template for the event.
% In the event causality identification task, KEPT leverages extra knowledge graphs to obtain background information and relational information for causal reasoning.$^{17}$

Path sampling provides another way to gather graph information in broader range for prompt design. GraphPrompt induces the topology structures of a protein graph $\mathcal{G}$ for the biomedical entity normalization task,$^{9}$ where each node in the protein-protein interaction graph represent a protein, and each edge represents the affiliations between two proteins. To predict the synonym of a given protein name, GraphPrompt constructs zero-order to second-order prompts of each candidate protein on $\mathcal{G}$. Here, the sampling function $g_{\mathrm{sample}}$ forms the sub-graph by collecting the $K$-hops neighbors of the center node, and the verbalizing function $g_{\mathrm{verb}}$ walks through paths from the center node to each $K$-hops neighbor and concatenates the textual definitions of nodes and edges on the path as output. 
%Suppose \emph{``CD115 (Human)"} is the synonym of the candidate term, where \emph{``CD115"} and \emph{``eukaryotic protein"} are 1-hop and 2-hops neighbors of ``CD115 (human)", GraphPrompt constructs its 0-order prompted input as \emph{``[MASK] is identical with \underline{CD115 (human)}"}, and its 1-order and 2-order prompted inputs are \emph{``[MASK] is a kind of \underline{human protein}"} and \emph{``[MASK] is a kind of \underline{human protein}, where is a kind of \underline{eukaryotic protein}"}.

\section{Continuous Graph Prompt}
\label{continuous_prompt}
Discrete templates describe the graph knowledge with words from a finite vocabulary set, which may not obtain the optimal representations of graph connectivity information. In contrast, continuous representations can capture the knowledge more accurately and comprehensively. The continues prompting function represents the graph knowledge as embeddings within continuous templates, which will then be concatenated with the embeddings of the input sample. Specifically, the continues prompting function could be written as:
\begin{equation}
    \label{eq: learn_prompt}
    \begin{aligned}
    \mathbf{x}' &= f_{\mathrm{prompt}}(\mathbf{x}; G, \mathbf{T}),
    \end{aligned}
\end{equation}
where $\mathbf{x}$ denotes the continuous representation of the input sample $x$, $\mathbf{x}'$ denotes the continuous representation to the prompted input $x^\prime$, and $\mathbf{T}\in\Re^{n\times d}$ denotes a trainable matrix. 
Here, $n$ and $d$ serve as hyper-parameters, and $\theta_\mathrm{prompt} = \{G, \mathbf{T}\}$ are the parameters of the prompting function. 
Different from the discrete template $\mathcal{T}$, the continuous template $\mathbf{T}$ is usually obtained by optimizing it with some training samples. In this subsection, we summarize research according to different levels of graph components they adopted in continue prompting functions, including graph irrelevant components, node attribute, and graph topology.

\subsection{Manual Prompt} % Graph Irrelevant, Manual Prompt Design
To help the pre-trained model use extra parameters to understand the downstream task, a simple approach is to concatenate the continuous template with input: % 
% a fundamental approach is to concatenate trainable embeddings to the input $\mathbf{x}$: 
\begin{equation}
\mathbf{x}^\prime = [\mathbf{x}; \mathbf{T}],
\label{LearnablePromptEmbedding}
\end{equation}
where $[\cdot;\cdot]$ is the concatenate operation. 
%we can obtain sufficient continuous prompts $\mathbf{T}$ by optimizing it with a small set of training samples $\mathcal{D}$.
Since the number of trainable parameters is small, we could sufficiently learn the continuous template $\mathbf{T}$ with a small set of training data $\mathcal{D}$.

This idea has been applied to pre-train graph neural networks, where task-specific prompt templates are learned to provide graph-level transformation on the downstream graphs during inference without tuning the parameters of the pre-trained GNN model.$^{18}$ 
%They point out that GPF can modify the node features and the graph structure implicitly. Their experiments show that GPF can achieve comparable performances as finetuning with a small amount ($\approx 0.1\%$) of tunable parameters. 
It points out that the learned continuous template $\mathbf{T}$, also known as the Graph Prompt Feature (GPF), can implicitly modify the node features and graph structure. 
The experiments demonstrate that this approach can achieve comparable performance to fine-tuning, with a minimal amount ($\approx 0.1\%$) of tunable parameters.

\subsection{Node-level Prompt} % Node Level Information
Learning prompt embeddings defined at Eq.~\eqref{LearnablePromptEmbedding} does not take into account the input $x$ or its corresponding node information in graph $G$. 
The most straightforward way to develop a graph-related prompt is to gather node $x$'s embeddings:
\begin{equation}
\mathbf{x}^\prime = [\mathbf{x};\mathbf{T}^\top g_{\mathrm{oh}}(x)],
\end{equation}
%where $\mathbf{T}\in\Re^{|\mathcal{V}|\times d}$ is the trainable parameters of the prompting function, and $g_{\mathrm{oh}}(v):\Re\rightarrow\{0,1\}^{|\mathcal{V}|}$ returns an one-hot vector indicating the position of node $x$ in the node embedding matrix $\mathbf{T}$.
where the number of rows at continuous matrix $\mathbf{T}$ is determined by the size of node set $\mathcal{V}$, and $g_{\mathrm{oh}}(x):\Re\rightarrow\{0,1\}^{|\mathcal{V}|}$ returns an one-hot vector indicating the node index of input $x$.
% $g_{\mathrm{oh}}(v):\Re\rightarrow\{0,1\}^{|\mathcal{V}|}$ returns an one-hot vector indicating the position of node $x$ in the node embedding matrix $\mathbf{T}$.

L. Lei et al. apply this strategy for explainable recommendation,$^{11}$ where they define graph $G$ as a bipartite graph with users and items.
%In their study, each user $u$ and item $i$ is considered as a node on the user-item interaction graph $G=(\mathcal{U}\cup\mathcal{I}, \mathcal{E})$, where edge $r_{u,i}\in\mathcal{E}$ indicates that the user $u$ purchased the item $i$. They also define $\mathbf{e}_u\in\mathbb{R}^d$ and $\mathbf{e}_i\in\mathbb{R}^d$ are the node embedding for the user $u$ and item $i$ respectively. 
Given a user-item pair as input $x=(u, i)$, the corresponding node embeddings $(\mathbf{e}_u, \mathbf{e}_i)$ are collected. They consider these node embeddings into the continuous prompts and input them into a pre-trained language model for generating explainable recommendations.

\subsection{Topology-level Prompt} % Topology Level Information
Encoding topology information with continuous representations motivates the diverse prompt designs. We consider several types of topological information in the graph prompt learning, which includes ontology, motifs and sub-graph structures. 
% \textcolor{red}{KX: Could we have a sentence to summarize the following technologies.}

\subsubsection{Ontology Embedding.}
Recall that a knowledge graph containing instances $\mathcal{V}$ and ontology $\mathcal{V}'$ as the node sets, a step-forward strategy to align the knowledge graph to the input $x\in\mathcal{V}$ is querying its ontology node $\widetilde{x}\in\mathcal{V}'$. 
Here, the ontology of a node represents the type of the node, introducing more human understanding to an entity. 
%with an embedding $\mathbf{z}_{\widetilde{x}}\in\Re^d$. 
For example, if \emph{``Bill Gates''} is the input node $x$, one of its ontology node $\widetilde{x}$ could be \emph{``Person''}. 
We formalize the continuous prompt indicated with ontology embedding as:
\begin{equation}
\begin{aligned}
\mathbf{x}^\prime &= [\mathbf{x};\mathbf{T}^\top g_{\mathrm{ont}}(x, G)], \\
\end{aligned}
\end{equation}
where $g_{\mathrm{ont}}$ is a function finding the ontology node $\widetilde{x}$ of the input node $x$ from the graph $G$, and the number of rows at continuous template $\mathbf{T}$ is given by the number of ontology.  

Promoting input $x$ by using its corresponding ontology embedding has been discussed in knowledge extraction under the zero/few-shot(s) settings. The reason is the rich prior ontology is critical to infer the semantic information of input sample.    
% The reason is that the rich ontology semantics and prior knowledge cannot be ignored in predicting the relation labels between pairs of entities. 
For example, if we know that the subject \emph{``Hamilton''} is a person and the object \emph{``British''} is a country, the prediction probabilities on the candidate relations irrelevant to person/location will be significantly weakened (e.g., \emph{``birth\_of\_organization''}  is an impossible relation). 
KnowPrompt first discusses combining the ontology knowledge of the subject-object pair for relation extraction tasks.$^{19}$
%KnowPrompt takes a subject-object pair $x=(x_s, x_o)$ as input. It finds out the corresponding ontology embeddings $(\mathbf{z}_{\widetilde{x}_s}, \mathbf{z}_{\widetilde{x}_o})$ of the inputs. Finally KnowPrompt designs its prompting function $f_{\mathrm{prompt}}((x_s, x_o), (\mathbf{z}_{\widetilde{x}_s}, \mathbf{z}_{\widetilde{x}_o}))$ as inserting the ontology embeddings to the input subject/object, and filling them into a pre-defined template. 
OntoPrompt extends this idea to a wider range of application scenarios,$^{15}$ including event extraction and knowledge graph completion. 
%OAG-BERT brings this idea to the pre-training stage so that the ontology embeddings $\mathbf{z}$ no longer require tuning in the downstream task.$^{20}$ 
Moreover, OAG-BERT brings this idea during the pre-training, so that the continuous template $\mathbf{T}$ could be obtained without downstream dataset.$^{20}$

%so that the estimation to the embeddings $\mathbf{T}$ does not require downstream samples.$^{20}$ 

\subsubsection{Motif Embedding.}
Furthermore, the frequently appearing sub-structures of a graph dataset typically have  special physical meanings and they are named as motif $M=(\widetilde{\mathcal{V}}, \widetilde{\mathcal{E}})$, where $\widetilde{\mathcal{V}}\subset\mathcal{V}$ and $\widetilde{\mathcal{E}}\subset\mathcal{E}$. For example, network motif mining yields insights in analyzing cellular signalling systems, and biochemical motifs are the functional elements of molecules. Thus, it is valuable to inject the informative motifs to input sample $x$ as prompt. 
Assuming we totally collect $m$ unique motifs from the target application,
% to the graph $G$, this idea is formatted as:
the continuous prompt is formatted as:
\begin{equation}
\begin{aligned}
\mathbf{x}^\prime &= [\mathbf{x};\mathbf{T}^\top g_{\mathbf{motif}}(x, G)], \\
\end{aligned}
\end{equation}
%where $\mathbf{Z}\in\Re^{m\times d}$ is the trainable motif embeddings, $g_{\mathrm{motif}}: G\rightarrow \{0,1\}^m$ is a motif detector returning a $m$-length binary vector. Elements of the binary vector is set as one if its corresponding motif appears within the input sample $x$.
where $g_{\mathrm{motif}}: G\rightarrow \{0,1\}^m$ is a motif detector returning a $m$-length binary vector, and $\mathbf{T}\in\Re^{m\times d}$ denotes the trainable embeddings for each motif structure.

The sub-structure analysis widely occurs in molecular representation learning, where input sample $x$ is a molecular graph and the motifs are some functional groups, such as a benzene ring. 
MolCPT is one of the classic method in this topic.$^{21}$
%MolCPT designs the motif detector $g_{\mathrm{motif}}$ as some rule-based methods, and the prompting function $f_{\mathrm{prompt}}$ as a self-attention module.  
Specifically, it first introduces human experts to define a set of key motif structures, and then designs motif detector $g_\mathrm{motif}$ to search them from the input graph.
To this end, the prompted input $\mathbf{x}^\prime$ contains the extra knowledge about motif structures to determine the underlying functions of physiological and biophysical molecules. 

\subsubsection{Message Passing.}
To fully explore the topology and neighbor features centered at input node $v$, % from the graph $G$, 
graph neural networks can be applied to gather information from its $k$-hop neighbors:
\begin{equation}
\label{message_passing}
\begin{aligned}
\mathbf{x}^\prime &= [\mathbf{x};g_{\mathrm{gnn}}(\widetilde{G};\theta_{\mathrm{gnn}})], \\
\end{aligned}
\end{equation}
where $\widetilde{G}$ refers to the $k$-hop sub-graph of input node $v$ extracted from the whole graph, and $g_{\mathrm{gnn}}$ is a graph neural network with trainable parameter $\theta_{\mathrm{gnn}}$.

Conversational recommendation task is one of the scenarios that needs the powerful representation ability of graph neural networks because of its complex user-item interactions. The node set $\mathcal{V}$ of graph $\mathcal{G}$ consists of the items and their attributes, and the edge between an item node and an attribute node indicates the belonging relation.  
UniCRS applies the message-passing-based prompt learning into this scenario.$^{22}$ 
Particularlly, during a conversation, when an user mentions an item $i\in\mathcal{V}$, UniCRS collects a $k$-hop sub-graph $\widetilde{G}_i\in G$,
and passes it to GNN encoder $g_{\mathrm{gnn}}$ to obtain a $d$-dimensional node embedding for each node at the sub-graph $\widetilde{G}_i$. 
Finally, these dynamically generated node embeddings are considered as the prompts to generate the final response for the user.

\subsubsection{Sub-Graph Pre-trained Embedding.} 
While the message-passing-based approach (Eq.~\ref{message_passing}) exhibits a strong ability to learn representations for sub-graph $\widetilde{G}$, training its parameter $\theta_\mathrm{gnn}$ from scratch necessitates a substantial amount of training data. 
%Learning the representations of graph sub-structures from scratch still requires a large number of training samples.  
To relax the requirement of training samples, one parametric-efficient way is first aligning a pre-trained model to initialize the representations of nodes:
%Aligning another pre-trained model to initialize the representations of nodes according to the node's attributes is a parameter-efficient way for this problem, so that :
\begin{equation}
\begin{aligned}
\label{pretrain_model_adaption}
\mathbf{x}^\prime &= [\mathbf{x};g_{\mathrm{ptm}}(\widetilde{\mathcal{G}})], \\
\end{aligned}
\end{equation}
where $g_{\mathrm{ptm}}$ is a pre-trained model with frozen weights, which generates embeddings for nodes from $\widetilde{\mathcal{G}}$.

Community question answering task aims to answer a use question by using the resources created by other users from the same community. 
This task perfectly fits in aligning pre-trained language models since the concepts from the community graph $G=(\mathcal{V}, \mathcal{R})$ are usually texts. 
%Community question answering task aims to answer a user question by aligning the resources created by other users from the same community, which perfectly fits in using pre-trained language models since the contents from the community graph $G=(\mathcal{V}, \mathcal{R})$ are usually texts. 
Here, the node set includes articles, comments, and questions, and the edges between the nodes are defined in some natural (e.g., a comment to an article) or weak-supervised (e.g., BM25 measures a pair of similar article and question) ways. 
Prefix-HeteroQA was the first attempt in this path.$^{23}$  
Given a question $x$ and its most similar question $q$ from the community, Prefix-HeteroQA searches $k$-hops neighbors of the question $q$ to form a sub-graph $\widetilde{G}$. 
% % 字数不够了
%This sub-graph could include other user's answers for question $q$, articles could be used to answer questions $q$, and the articles comments. 
Eq.~\eqref{pretrain_model_adaption} is finally applied to obtain the embeddings of each document from the sub-graph as prompts to enhance the input question $x$.

\section{Applications}
% 1. 把RecSys子章节拆分到Graph-Learning和NLP里面
% 2. 把NLP里面再按照NLP任务细分
%We summarize and introduce the applications of graph prompting methods in various domains. 
% % 字数不够
% In this section, we summarize the graph prompting methods according to their applications, including graph learning and natural language processing. 

\subsection{Graph Learning}
\subsubsection{Homogeneous Graphs.} 
Homogeneous graphs are typically encountered in pure graph learning tasks, such as node classification, link prediction, or graph classification tasks.
%They usually follow the standard supervised training process on benchmark datasets (e.g., Cora, Reddit, PPI) but empowered with a pre-trained graph model. To better introduce the pre-training knowledge, these methods apply prompts to reformulate the downstream task looking similar to the pretext one, which can reduce the training objective gap between the constructed pretext and dedicated downstream tasks.$^{24,27,28}$ 
In contrast with the traditional supervised training process, these methods align pre-trained graph models with prompt learning. 
Specifically, they design prompts to reformulate the downstream task (e.g., node/graph classification) as the pretext task (e.g., link prediction) for training the pre-trained models, which can reduce the training objective gap between the constructed pretext and dedicated downstream tasks.$^{24,27,28}$

\subsubsection{Heterogeneous Graphs.} 
Heterogeneous graphs are prevalent in many real-world applications, including knowledge graphs, user-item interaction graphs, and social networks.  
Knowledge extraction is one of the most common applications for knowledge graphs, which extracts entities from a piece of input text and predicts the relations among these entities. 
Graph prompting methods align additional knowledge, typically stored in another knowledge graph, with words and phrases to the input raw text. 
For example, the phrase ``Bill Gates'' belongs to the ``Person'' ontology and ``Microsoft'' is an entity of the ``Company'' ontology.$^{15}$ 
These additional information help the models make more precise prediction.$^{12, 15, 19}$ 
In addition, recommender systems also attract studies in graph prompt learning due to several reasons: (1) Users, items, and their interactions naturally form a graph, where the node set is the users and items, and the edge set can be various types of user-items interactions (e.g., shopping, clicking, commenting, visiting, favoring, and so on).
(2) Many components in recommendation systems (e.g., names of users and items, authors of items, categories of items) can be represented with natural language, simplifying the designs of discriminate prompts. 
More specifically, graph prompt learning has been applied to enhance the recommendation systems for various purposes, such as interpretability,$^{7,11}$ multi-tasks learning,$^{7}$ sequential recommendation,$^{13,14,26}$ and conversational recommendation.$^{22}$ 

\subsection{Natural Language Processing} 
\subsubsection{Text Understanding.} 
Text understanding could refer to various natural language tasks, such as sentiment analysis, topic classification, named entity recognition, and retrieve-based question answering. 
Traditionally, these tasks are usually addressed by using the input text, as it can be challenging to represent the context and background knowledge contained in other formats (such as knowledge graphs) using language models. 
However, recent studies found that prompt learning could overcome this challenge without redesigning a language model. 
Many of them reformat the extra knowledge as part of the textual context and append them as the additional information to the original input text during the inference stage.$^{23, 25}$ 
Other research incorporates graph prompting during the pre-training of the language model, enabling the models to inherently integrate graph information. 
One example of this is OAG-BERT,$^{20}$ which is designed for the topic classification task over the academic articles. 
During its pre-training, the input content is assigned both word embeddings and category embeddings, such as title, author names, and abstract, which help OAG-BERT to better understand the context's ontology. 
%Text understanding typically refers to the text classification task, whether it is a single word or a longer piece of text.  
%By providing additional context and background knowledge, these methods can improve the accuracy of classification models. 
%To take advantage of this approach, some studies incorporate graph prompting during the pre-training stage of a model, enabling it to learn how to integrate graph information directly into its understanding of text.$^{20}$ 
%But more studies brings these graph knowledge during the inference, which has no need to pre-train a new language model.$^{23, 25}$

\subsubsection{Text Generation.}
Text generation is much challenging than the text understanding, as it requires the model to generate choose words from the whole vocabulary set to represent its understanding and responses to the input text. 
Similar to the text understanding tasks, graph prompting methods could provide more context and additional knowledge to the generator as a better guidance to the generation process. 
For example, H. Liu and Y. Qin propose to enhance the context of an input question for better answer generation by aligning a community content graph,$^{23}$ where the content graph consists of articles, comments, questions, and answers from other users on this community. 
Although most of these graph prompting methods are introduced during the inference,$^{11, 14, 22, 23, 26}$ there are still some researchers proposed to prompting graph knowledge started from the pre-training stage.$^{7}$

\section{Challenges}
%Although incorporating graph-structured knowledge in prompt learning has been proven as a sufficient way to improve the vanilla prompt learning method, some emergency problems still need to be solved. 
% %In this section, we discuss in detail some key challenges related to incorporating graph-structured knowledge into prompt learning.

\subsubsection{Pre-training GNN Models.}
In light of the huge success of pre-trained models in text mining, many tasks in graph related domains (e.g., knowledge graph extraction and recommendation) are transformed into NLP problems. 
However, a GNN-based model instead of the language model could better encode the structure knowledge of graph database. 
% However, it is much more reasonable to apply a GNN-based model instead of a language model as the backbone pre-trained model in those graph-related domains, where the structures of the input samples are much more important. 
Recently, researchers have begun to design the more general graph pre-training objectives to improve the generalization of pre-trained GNNs for better prompt learning.$^{27,28,29}$

% \subsubsection{Pre-trained Model Selection.}
% %Since many graph prompt learning methods heavily rely on the pre-trained language models, this emergency problem in NLP with prompt learning is also raised in the topic of graph-based prompt learning. 
% %To the best of our knowledge, it still lacks studies in predicting pre-trained model performance with prompt learning, especially under the zero-shot setting.
% Prompt learning is built on top of large pre-trained models. 
% In recent years, the rapid advancement of pre-training models has given rise to a new challenge in applying prompt learning methods: selecting the most appropriate pre-training model for a given task. 
% To the best of our knowledge, it still lacks studies in predicting the pre-trained model performance, especially under the zero-shot setting. 
% As an increasing number of open-source models emerge on public platforms, it becomes increasingly challenging to choose the best pre-training model when data is not available. 

\subsubsection{Knowledge Injection Answer Mapping.}
Current methods that rely on manually designed rules incorporate structured information into answer mapping functions. 
However, this approach introduces a strong bias from the designers and results in low efficiency. 
Some attempts have been made in the field of knowledge extraction, where they inject extra knowledge graphs to automatically optimize the answer mapping function.$^{19,25}$ 
We encourage future works to broaden this idea to wider applications.

\subsubsection{Non-Generative Answer Mapping.}
Different from the pre-training objectives in NLP and CV, pre-training objectives of graph neural networks usually refer to discriminative tasks. 
The form of downstream tasks that can be supported is constrained by the output of these discriminative pre-training tasks. 
For example, predicting whether an edge is exist or not is a typical pre-training objective to pre-train graph neural networks, where the output is a probability ranged between 0 to 1. 
To compare, the output space of pre-trained NLP models usually is the vocabulary set, while that of pre-trained CV models could be an image. 
Therefore, it is challenging to design sufficient answer mapping functions that can release the potentials of pre-trained graph neural networks.

\subsubsection{Explanation and Fairness.}
%Since graph-structured data is natural to organize the explicit relationships between nodes, 
%  easier to understand by humans,
%it is desirable to further exploit such structured side information to develop explainable and fair systems. 
Given the inherent nature of graph-structured data in capturing explicit relationships between nodes, there is a growing interest in leveraging this structured side information to develop explainable and fair systems. For instance, in the context of biomedical entity normalization, the use of a biomedical entities graph allows for a clear pathway in the entity normalization process, facilitating human experts in validating the outputs of the model.$^9$ 
Furthermore, other applications driven by the need for explanation can also benefit from incorporating additional graph-structured data to enhance their performance through prompt learning, such as developing explainable recommender systems.$^{7,11}$
% % 字数不够了
%. Notably, researchers have explored the application of graph prompt learning in developing explainable recommender systems.$^{7,11}$
%Some researchers have developed explainable recommender systems by using graph prompt learning.$^{7,11}$
%We believe they will inspire and motivate future research to examine interpretable graph prompt learning methods in other areas, domains, and applications.

\section{Conclusion}
In this survey, we have explored the emerging learning paradigm of ``pre-train, prompt, predict'' from graph-related learning scenarios. Our analysis highlights the essential role of graphs as structured knowledge repositories that enable adaptive prompt templates and generalization of pre-trained models to downstream scenarios with limited labeled data. We have identified two types of graph prompting designs: discrete and continuous, each with their own strengths and limitations. Our survey has demonstrated the applications of these techniques in graph representation learning and natural language processing. We conclude that graph prompting functions have significant potential to enhance the generalization of graph applications and personalize prompt templates in NLP. However, we have also identified several challenges. Overall, our study provides insights and directions for future research in this exciting area.

\newpage
\def\refname{REFERENCES}
\vspace*{-8pt}

\begin{IEEEbiography}{Xuansheng Wu} currently is a second-year Ph.D. student in the School of Computing at the University of Georgia, Athens, GA, USA. His research interests broadly cover Natural Language Processing, Recommendation Systems, and Representation Learning. Contact him at xuansheng.wu@uga.edu.
\end{IEEEbiography}

\begin{IEEEbiography}{Kaixiong Zhou}{\,}received the BS degree in electrical engineering and information science from Sun
Yat-Sen University, and the MS degree in electrical
engineering and information science from the University of Science and Technology of China. He is
currently working toward the PhD degree with the
Department of Computer Science, Rice University.
His research interests include large-scale graph machine learning and its applications in science. Contact him at Kaixiong.Zhou@rice.edu.\vadjust{\vfill\pagebreak}
\end{IEEEbiography}

\begin{IEEEbiography}{Mingchen Sun}{\,} is a Master student at Jilin University, China. His interests include Meta Learning, Graph Data Mining, and Domain Generalization.\vspace*{8pt}
\end{IEEEbiography}

\begin{IEEEbiography}{Xin Wang} {\,} Ph.D., associate professor at Jilin University, China. He is also a senior member of CCF. His main research interests include machine learning, information retrieval, and social computing.
\end{IEEEbiography}

\begin{IEEEbiography}{Ninghao Liu} is an assistant professor in the School of Computing at the University of Georgia, Athens, GA, USA. His research interests are Explainable AI (XAI), Natural Language Processing, and Graph Mining. Contact him at ninghao.liu@uga.edu.
    
\end{IEEEbiography}

\end{document}